%% file: main.tex
\documentclass[10pt,twocolumn,letterpaper]{article}

\usepackage[pagenumbers]{cvpr} 

\input{preamble}
\definecolor{cvprblue}{rgb}{0.21,0.49,0.74}
\usepackage[pagebackref,breaklinks,colorlinks,allcolors=cvprblue]{hyperref}
\usepackage{multirow}
\usepackage{xcolor}

\def\paperID{948} 
\def\confName{CVPR}
\def\confYear{2026}

\title{COOPER: A Unified Model for Cooperative Perception and Reasoning \\ in Spatial Intelligence}

\author{
Zefeng Zhang$^{1,2}$\thanks{Equal contribution.} \quad
Xiangzhao Hao$^{3}$\footnotemark[1] \quad
Hengzhu Tang$^{4}$ \quad
Zhenyu Zhang$^{4}$ \quad
Jiawei Sheng$^{1,2}$ \\
Xiaodong Li$^{1,2}$ \quad
Zhenyang Li$^{4}$ \quad
Li Gao$^{4}$ \quad
Daiting Shi$^{4}$ \quad
Dawei Yin$^{4}$ \quad
Tingwen Liu$^{1,2}$ \\
$^{1}$Institute of Information Engineering, Chinese Academy of Sciences \\
$^{2}$School of Cyber Security, University of Chinese Academy of Sciences \\
$^{3}$Institute of Automation, Chinese Academy of Sciences \\
$^{4}$Baidu Inc. \\
{\tt\small zhangzefeng@iie.ac.cn}
}
\begin{document}
\maketitle
\input{sec/0_abstract}    
\input{sec/1_intro}
\input{sec/2_preliminaries}
\input{sec/3_method}
\input{sec/4_experiments}
\input{sec/5_related}
\input{sec/6_conclusion}


{
    \small
    \bibliographystyle{ieeenat_fullname}
    \bibliography{main}
}

\input{sec/X_suppl}

\end{document}

%% file: sec/0_abstract.tex
\begin{abstract}
Visual Spatial Reasoning is crucial for enabling Multimodal Large Language Models (MLLMs) to understand object properties and spatial relationships, yet current models still struggle with 3D-aware reasoning. Existing approaches typically enhance either perception, by augmenting RGB inputs with auxiliary modalities such as depth and segmentation, or reasoning, by training on spatial VQA datasets and applying reinforcement learning, and thus treat these two aspects in isolation. 
In this work, we investigate whether a unified MLLM can develop an intrinsic ability to enhance spatial perception and, through adaptive interleaved reasoning, achieve stronger spatial intelligence.
We propose \textbf{COOPER}, a unified MLLM that leverages depth and segmentation as auxiliary modalities and is trained in two stages to acquire auxiliary modality generation and adaptive, interleaved reasoning capabilities. 
COOPER achieves an average \textbf{6.91\%} improvement in spatial reasoning while maintaining general performance. Moreover, even a variant trained only for auxiliary modality generation attains a \textbf{7.92\%} gain on distance and size estimation, suggesting that learning to generate auxiliary modalities helps internalize spatial knowledge and strengthen spatial understanding.
\end{abstract}

%% file: sec/1_intro.tex
\section{Introduction}
\label{sec:intro}

Visual Spatial Reasoning~\citep{yang2025thinking} investigates how models perceive, understand, and reason about object properties and spatial relationships.
It represents a fundamental step toward enabling vision-language models to achieve human-level intelligence, while also serving as a cornerstone for numerous downstream applications in robotics~\citep{driess2023palm, brohan2022rt, zitkovich2023rt, o2024open}, autonomous driving~\citep{tian2024drivevlm} and AR/VR~\citep{chandrasegaran2024hourvideo, grauman2022ego4d, mangalam2023egoschema} fileds.
Despite the recent advancements of Multimodal Large Language Models (MLLMs)~\citep{bai2025qwen2, wang2025internvl3, hurst2024gpt, comanici2025gemini, guo2025seed1}, their compositional spatial reasoning capabilities remain limited and fall far behind human-level performance~\citep{yang2025thinking, yu2025far, li2025sti}.

\begin{figure}
\centering
\includegraphics[width=0.48 \textwidth,height=0.42\textwidth]{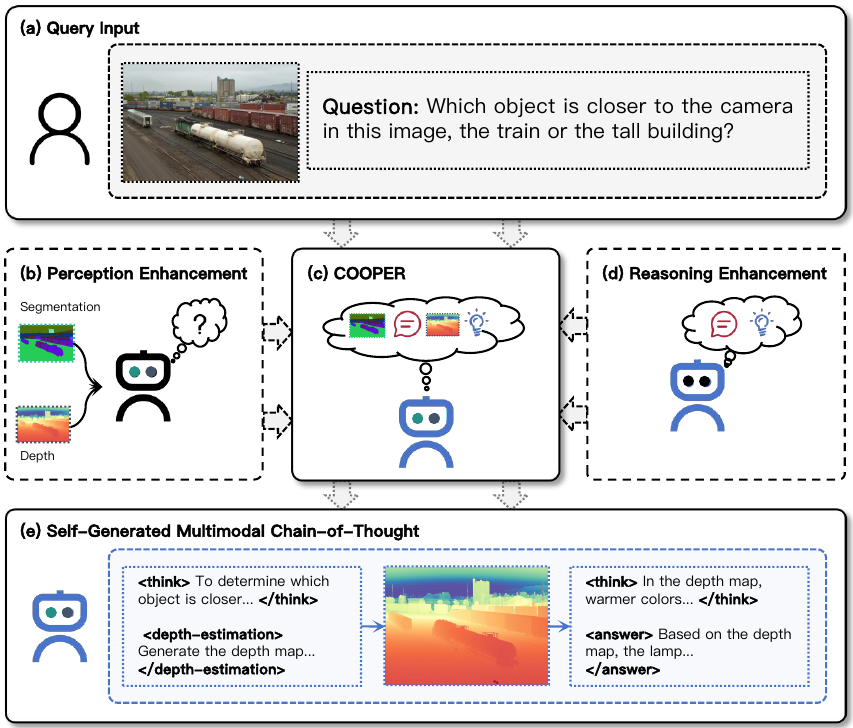}
\caption{Comparison of three paradigms. (a) Query input: visual and the corresponding textual information. (b) \textbf{Perception enhancement}: augment the model with auxiliary modalities (e.g., depth, segmentation). (c) \textbf{COOPER}: a single model endowed with both capabilities that adaptively schedules when to perceive and when to reason during execution. (d) \textbf{Reasoning enhancement}: strengthen spatial reasoning via textual chain-of-thought. (e) Self-generated Multimodal CoT: an interleaved vision–language CoT generated by the unified reasoner.} 
\label{fig:motivation}
\end{figure}

Conventional MLLMs are predominantly trained on 2D image–text pairs and therefore tend to exhibit limited 3D awareness: they often struggle to infer geometry, depth, and object boundaries from raw pixels. To alleviate this limitation and strengthen spatial understanding, existing efforts mainly follow two lines of work:
First, \textit{perception enhancement} methods introduce auxiliary modalities such as depth maps, semantic segmentation, and 3D point clouds to strengthen the model’s spatial perception~\citep{cheng2024spatialrgpt, zhou2025roborefer, liu2025ssr}. By enriching low- and mid-level visual features through fixed perception pipelines, these approaches can indeed improve the extraction of geometric cues. However, they typically offer only marginal gains in high-level spatial reasoning and downstream decision-making.
Second, \textit{reasoning enhancement} typically improves spatial reasoning by strengthening the model’s ability to decompose and analyze complex problems through textual chain-of-thought.
For example, recent works construct spatial VQA datasets with annotated reasoning trajectories~\citep{ray2024sat, liu2025spatialcot} and employ reinforcement learning to refine reasoning policies~\citep{ai2025m2, ouyang2025spacer, ma2025spatialreasoner}. 
However, these methods perform spatial reasoning solely from raw 2D images, without explicit 3D grounding, which leads to brittle and inconsistent spatial understanding.
Taken together, perception and reasoning constitute two interdependent pillars of spatial intelligence: without precise geometric perception, high-level planning is prone to error; conversely, without robust reasoning, even the most accurate geometric perception struggles to support complex tasks.
This naturally leads to the question: \textbf{Can a single model unify perception and reasoning in a cooperative way to achieve stronger spatial intelligence?}

Unified MLLMs~\citep{deng2025emerging}, with their strong multimodal understanding and generation capabilities over both images and text, offer a promising direction for the question. However, bridging perception and reasoning within a single unified model is far from trivial:
First, \textit{it is non-trivial for a unified MLLM to generate non-RGB auxiliary modalities that are informative for spatial reasoning}. In principle, spatial perception can be greatly facilitated by explicit auxiliary modalities such as depth maps and segmentation masks. However, unified MLLMs are optimized to produce photorealistic RGB images, and are not equipped to generate structured non-RGB auxiliary modalities. This mismatch between what perception enhancement typically relies on and what unified MLLMs currently generate means that the model cannot readily supply the explicit spatial cues that would support more robust spatial reasoning.
Second, \textit{adaptive, interleaved spatial reasoning remains particularly challenging for unified MLLMs}. Existing unified MLLM frameworks generally follow predefined pipelines for image or text generation and seldom allow the model to autonomously decide when to generate images versus text. Even if the model were capable of producing multiple visual modalities, the lack of a flexible control mechanism for deciding \emph{when}, \emph{what}, and \emph{how} to generate these modalities limits its ability to tightly couple perception and reasoning, thereby constraining its performance on complex spatial reasoning tasks.

To address these challenges, we propose \textbf{COOPER}, a unified MLLM that cooperatively unifies perception and reasoning to achieve stronger spatial intelligence. 
Specifically, It unfolds in two core stages.
(1) \textit{Auxiliary modality generation.} 
To equip the model with the ability to generate non-RGB auxiliary modalities, we aggregate open-source depth and segmentation datasets and convert depth maps and segmentation masks into RGB pseudo-images, so that they are compatible with the flow matching-based training and inference pipeline of unified MLLMs.
Within this shared RGB space, the model jointly learns depth and segmentation, while a designated control token during inference dynamically selects the appropriate decoder to reconstruct depth or segmentation maps for downstream spatial reasoning.
(2) \textit{Adaptive reasoning.}
To enable the model to perform adaptive, interleaved spatial reasoning, we first undergoes supervised fine tuning on GPT-4o curated data to acquire task- and context-aware capability selection. Reinforcement learning then refines this policy with a Cooperative Perception–Reasoning reward (CPR reward) to further optimize spatial reasoning behavior by balancing exploration and exploitation.

We evaluate COOPER on three spatial perception and reasoning benchmarks, where it improves average spatial reasoning performance by \textbf{6.91\%} over the base model and significantly surpasses Perception Enhancement and Reasoning Enhancement baselines. 
On two general multimodal benchmarks, it also yields a modest overall gain of 4.47\%.
Further experiments show that even a variant trained only for auxiliary modality generation (without reasoning supervision) achieves a \textbf{7.92\%} improvement on distance and size estimation, indicating that learning to generate auxiliary modalities helps internalize spatial knowledge and strengthen spatial understanding.

The main contributions are summarized as follows:
\textit{(1) New paradigm.}
We propose a interleaved reasoning paradigm for spatial intelligence that cooperatively unifies perception and reasoning, enabling the model to both generate and exploit them for spatial reasoning.
\textit{(2) Training pipeline.}
We design a fine-grained training pipeline in which unified MLLMs first learn to generate auxiliary modalities, and are then optimized with RL using the CPR reward to acquire adaptive, interleaved reasoning abilities.
\textit{(3) Empirical validation.}
We conduct extensive experiments on spatial reasoning and general multimodal benchmarks, showing clear gains in spatial intelligence while maintaining broad multimodal capabilities, validated by comprehensive ablations and analyses.


%% file: sec/2_preliminaries.tex
\section{Preliminaries}
\label{sec:preliminaries}

\subsection{Unified Multimodal Large Language Model}
\label{sec:pre_unified_mllms}
This section briefly reviews the core capabilities and research focus of unified MLLMs and introduces the BAGEL framework used as our backbone; other variants are discussed in Related Works. By a unified MLLM, we mean a single model that jointly performs multimodal understanding and generation, with most existing work focusing on image and text. BAGEL adopts a Mixture-of-Transformer-Experts (MoT) architecture with two transformer experts—one for multimodal understanding and one for multimodal generation—and correspondingly uses separate understanding- and generation-oriented visual encoders. The two experts operate on a shared token sequence via common self-attention at every layer. For text, BAGEL follows the standard next-token prediction paradigm, while for visual tokens it employs Rectified Flow~\citep{lipman2022flow, liu2022flow, esser2024scaling}, in line with state-of-the-art visual generation practice.

\paragraph{Visual understanding.}
BAGEL leverage a ViT encoder to convert raw pixels into tokens. Concretely, it adopts SigLIP2-so400m/14~\citep{tschannen2025siglip} with a fixed $384$-resolution as the initialization of the ViT encoder. Building upon this, BAGEL interpolate the positional embeddings and set $980\times980$ as the maximum input size, and further integrate NaViT~\citep{dehghani2023patch} to enable processing images at their native aspect ratios. A two-layer MLP connector is used to match the feature dimension of the ViT tokens to the hidden states of the language model, thereby allowing the understanding expert to consume visual tokens in a unified token space.

\paragraph{Visual generation.}
BAGEL operates in latent space using a Rectified Flow formulation. Given an RGB image $\mathbf{x}\in\mathbb{R}^{w\times h\times 3}$ and a condition $c$ (e.g., a text prompt), a VAE encoder produces a target latent $\mathbf{z}_1\in\mathbb{R}^{d}$. A source latent $\mathbf{z}_0\sim\mathcal{N}(0,I)$ and a time $t\sim\mathcal{U}[0,1]$ are sampled, and a conditional linear path (linear flow/bridge) is defined as
\begin{equation}
\small
\label{eq:noise}
\begin{aligned}
\mathbf{z}_t=(1-t)\,\mathbf{z}_0+t\,\mathbf{z}_1,\qquad t\in[0,1].
\end{aligned}
\end{equation}
The ground-truth conditional velocity along this path is constant,
$\mathbf{u}^\star(\mathbf{z}_t,t\mid \mathbf{z}_0,\mathbf{z}_1)=\mathbf{z}_1-\mathbf{z}_0$.
The velocity network $v_\theta(\cdot)$ is trained via flow matching:
\begin{equation}
\small
\label{eq:diffusion_loss}
\begin{aligned}
\mathcal{L}_{\text{FM}}
=\mathbb{E}_{\mathbf{z}_0\sim\mathcal{N}(0,I),\,(\mathbf{z}_1,c),\,t\sim\mathcal{U}[0,1]}
\big\|\,v_\theta(\mathbf{z}_t, t, c)-(\mathbf{z}_1-\mathbf{z}_0)\big\|_2^2.
\end{aligned}
\end{equation}
At inference time, given condition $c$, we draw an initial latent $\mathbf{z}_0\sim\mathcal{N}(0,I)$ and solve the ODE induced by the learned velocity field
\begin{equation}
\small
\label{eq:denoise}
\begin{aligned}
\frac{d\mathbf{z}_t}{dt}=v_\theta(\mathbf{z}_t,t,c),\qquad t\in[0,1],
\end{aligned}
\end{equation}
using a numerical solver (e.g., Euler/Heun) with $T$ steps to obtain $\hat{\mathbf{z}}_1$. Finally, the VAE decoder maps $\hat{\mathbf{z}}_1$ back to RGB space to produce the image $\hat{\mathbf{x}}\in\mathbb{R}^{w\times h\times 3}$.

\subsection{Group Relative Policy Optimization}
The process begins with an input $q$, for which the policy $\pi_{\theta}$ samples $N$ responses $G = \{o_1,\ldots,o_N\}$, each evaluated by a composite reward function $R$ to yield $r_i=R(q,o_i)$. GRPO then computes a group-relative advantage $A_i$ for each response by normalizing its reward with respect to the statistics of the entire group, and updates the policy:
\begin{equation}
\small
\label{eq:advantages}
\begin{aligned}
A_i = \frac{r_i - \text{mean}\{r_1, \ldots, r_N\}}{\text{std}\{r_1, \ldots, r_N\}}.
\end{aligned}
\end{equation}
The policy is then updated by maximizing a clipped surrogate objective that favors responses with higher relative advantages while preventing overly large policy updates, ensuring stable training. The full objective is:
\begin{equation}
\small
\label{eq:grpo}
\begin{aligned}
J_{\mathrm{GRPO}}(\theta)
&= \mathbb{E}_{o_i \in G}\Biggl[\frac{1}{N} \sum_{i=1}^{N} \min\bigl(s_i A_i,\; \\ &\operatorname{clip}(s_i,1-\varepsilon,1+\varepsilon)A_i\bigr)
- \beta \,\mathbb{D}_{\mathrm{KL}}\!\left(\pi_\theta \,\Vert\, \pi_{\mathrm{ref}}\right)
\Biggr].
\end{aligned}
\end{equation}
Here, \( s_i = \dfrac{\pi_\theta(o_i \mid q)}{\pi_{\theta_{\mathrm{old}}}(o_i \mid q)} \)
is the importance sampling ratio that measures the change between the new policy
\( \pi_\theta \) and the old policy \( \pi_{\theta_{\mathrm{old}}} \) used to generate the samples.
The Kullback–Leibler (KL) divergence penalty,
\( \mathbb{D}_{\mathrm{KL}}\!\left(\pi_\theta \,\Vert\, \pi_{\mathrm{ref}}\right) \),
regularizes the policy update by penalizing large deviations from a reference policy \( \pi_{\mathrm{ref}} \).

%% file: sec/3_method.tex
\section{Method}
\label{sec:method}


In this section, we present the overall training pipeline of COOPER, which consists of two stages.
Section~\ref{sec:auxiliary_odality_eneration} introduces the first stage, \textit{Auxiliary Modality Generation}, where we endow the model with the ability to generate auxiliary modalities by mapping them into RGB space and training them directly with the original generative loss.
Section~\ref{sec:adaptive_reasoning} then describes the second stage, \textit{Adaptive Interleaved Reasoning}, where to equip the model with adaptive, interleaved reasoning capabilities, we adopt an SFT+RL paradigm and design a Cooperative Perception–Reasoning reward (CPR reward) that balances exploration and exploitation, enabling the model to learn such reasoning behavior.

\begin{figure*}
\centering
\includegraphics[width=1\textwidth]{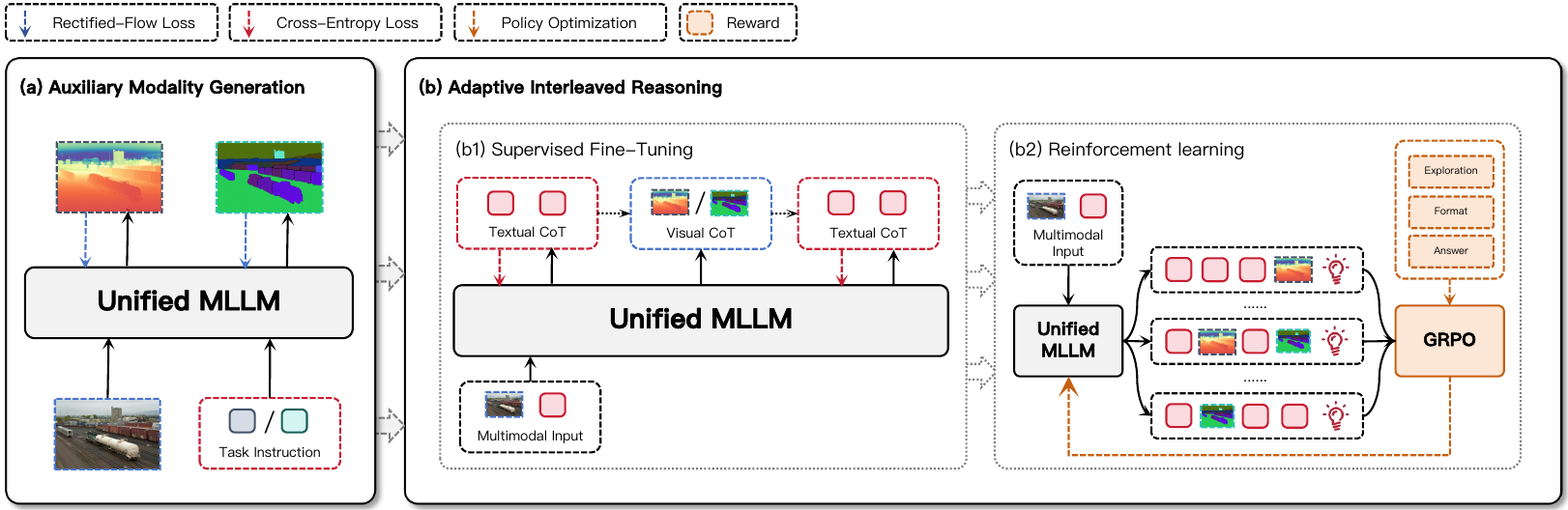}
\caption{\textbf{Method details.} The method consists of two stages:
\textit{(a) Auxiliary Modality Generation.} To equip the model with the ability to generate different types of auxiliary modalities, we convert all auxiliary-modality data into the RGB space and train the model to generate these modalities using the original image generation training pipeline.
\textit{(b) Adaptive Interleaved Reasoning.} Building on the model with auxiliary modality generation capability, we construct a balanced dataset and first apply supervised fine-tuning (SFT) to endow the model with basic interleaved reasoning. We then further enhance its reasoning and generalization ability using the CPR reward and GRPO.
} 
\label{fig:model}
\end{figure*}

\subsection{Auxiliary Modality Generation}
\label{sec:auxiliary_odality_eneration}
In this section, we equip BAGEL with specialist capabilities that enable it to generate auxiliary modalities to support its own spatial reasoning. We focus on two widely used signals: depth estimation (geometric) and segmentation (semantic). To fully exploit the model’s native capacity, we inject these capabilities without changing its architecture or training objective. We first show how to map task-specific ground truths into the RGB space to align with BAGEL’s flow matching-based training/inference pipeline, and then present the full Capability Infusion training and inference procedures, as illustrated in Figure~\ref{fig:model}~(a).

\paragraph{Representing auxiliary modalities in RGB space.} 
As described in Section~\ref{sec:pre_unified_mllms}, the image generator operates in RGB space, while depth and segmentation labels are single-channel maps, making them incompatible with RGB-based training and inference.
(1) For segmentation, we assign distinct RGB colors to different instances, turning the integer mask into an RGB label image that can be directly used in BAGEL’s image generation pipeline.
(2) For depth estimation, the diffusion model’s VAE expects both inputs and outputs to lie in $[-1, 1]$. Accordingly, following the Marigold~\citep{ke2024repurposing}, we first replicate the ground-truth depth map to three channels $\mathbf{x} \in \mathbb{R}^{w \times h \times 3}$, then apply an affine transformation to map its values to $\tilde{\mathbf{x}}\in [-1,1]$ that aligns with the VAE’s numerical range:
\begin{equation}
\small
\label{eq:affine_transformation}
\begin{aligned}
\tilde{\mathbf{x}} = (\frac{\mathbf{x} - \mathbf{x}_{2}}{\mathbf{x}_{98} - \mathbf{x}_{2}} - 0.5) \times 2,
\end{aligned}
\end{equation}
where $\mathbf{x}_{2}$ and $\mathbf{x}_{98}$ correspond to the $2\%$ and $98\%$ percentiles of individual depth maps. This normalization allows us to focus on pure affine-invariant depth estimation. During inference, we take the channel-wise mean of the decoder’s three-channel output to obtain the final depth prediction.

\paragraph{Training and inference.}
During training, we continue to optimize the model using the flow matching loss in Equation~\ref{eq:diffusion_loss}. In both training and inference, the conditioning signal provided to the model consists of the input image and the task-specific prompt. We use the \texttt{<depth-estimation>...</depth-estimation>} for the depth estimation task prompt, and \texttt{<segmentation>...</segmentation>} for the segmentation task prompt.

\subsection{Adaptive Interleaved Reasoning}
\label{sec:adaptive_reasoning}
Even with multiple auxiliary modalities, existing models still follow fixed image/text pipelines and cannot adaptively select the right capability. To enable adaptive reasoning, we use a SFT+RL framework: SFT on interleaved vision–language CoT data, followed by reinforcement learning with a tailored reward.

\paragraph{Data construction.}
High-quality data filtering and balancing are critical for both Supervised Fine-Tuning (SFT) and reinforcement learning (RL)~\citep{su2025pixel}. Accordingly, this section details how we construct and curate the training datasets for these two stages.
We first collect the SAT VQA dataset~\citep{ray2024sat} for spatial reasoning and the general QA dataset TACO~\citep{ma2024taco} as seed data. We then evaluate the original BAGEL model on each dataset in two sampling rounds: in each round, we draw $k=8$ responses per question. The first round uses only the raw input and computes the average accuracy, denoted $\mathrm{acc}_{\text{raw}}$; the second round augments the raw input with depth and segmentation maps and computes the corresponding average accuracy, denoted $\mathrm{acc}_{\text{aux}}$. For both rounds, the sampling temperature is set to $\tau=1.0$.
We first discard samples with $\mathrm{acc}_{\text{raw}}\in\{0,1\}$ to avoid overly hard or trivial questions that yield zero advantage during RL training, thereby weakening the learning signal.
We then partition the remaining data using the accuracy gap and a threshold $\lambda=0.375\,(=3/8)$: 
\begin{equation}
\small
\label{eq:visual_gain}
\begin{aligned}
\text{gain} =
\begin{cases}
\text{positive}, & \mathrm{acc}_{\text{aux}}-\mathrm{acc}_{\text{raw}}>\lambda,\\[2pt]
\text{negative}, & \mathrm{acc}_{\text{raw}}-\mathrm{acc}_{\text{aux}}>\lambda,\\[2pt]
\text{boundary}, & \text{otherwise},
\end{cases}
\end{aligned}
\end{equation}
where \texttt{boundary} indicating no significant effect from the auxiliary visual modalities.

Next, we randomly subsample the boundary-type examples to control the dataset size, and split them evenly: half for SFT and the other half for RL.
To construct SFT data in an interleaved vision–language CoT format, we use GPT-4o as the agent. To ground the reflections in the model’s actual capabilities, we employ the BAGEL after thaining with Section~\ref{sec:auxiliary_odality_eneration} as callable tools for depth estimation and segmentation, supplying visual signals and provisional answers for each query. When BAGEL’s generated images contain misleading cues, GPT-4o produces explicit reflections in the CoT to correct them. We then retain only the examples with correct final answers as the SFT dataset.

\paragraph{Supervised fine-tuning.}
After obtaining the synthetic interleaved vision–language CoT data, we conduct SFT with a cross-entropy objective. Concretely, we supervise only the textual reasoning and answer tokens in the CoT; since the visual content is generated by the model itself, optimizing it would introduce additional noise and target drift, so we do not optimize the visual outputs. The detailed SFT training pipeline is provided in Figure~\ref{fig:model}~(b1).

\paragraph{Reinforcement learning with CPR reward.}
After SFT, the model can adaptively activate generation capabilities and reason, but SFT primarily teaches pattern memorization and does not generalize well.
Accordingly, we further train the warm-started model with standard GRPO (described in Euation~\ref{eq:grpo}) using a composite \emph{Cooperative Perception–Reasoning reward} (CPR reward) to enhance its adaptive reasoning and generalization capabilities. Specifically, our CPR reward consists of three components: an \emph{answer reward} $r_a$, a \emph{format reward} $r_f$, and an \emph{exploration-guided reward} $r_e$, and the total reward is given by
\[
\small
R_{CPR} \;=\; r_a + r_f + r_e.
\]
(1) For the \textit{answer reward},
we use a rule-based checker to verify the final answer; if correct $r_a=1.0$, otherwise $r_a=0.0$.
(2) For the \textit{format reward},
the interleaved CoT takes the form $\{t_1, v_2, \ldots, t_N\}$, where $t_i$ are textual segments and $v_i$ are visual segments.
If all $t_i$ for $i\in\{1,\ldots,N-1\}$ match thinking-generation pattern
and the final $t_N$ matches thinking-answer pattern
then $r_f=1.0$; otherwise $r_f=0.0$, you can find the details of these patterns in the Appendix.
(3) For the \textit{exploration-guided reward}, we use offline visual-gain labels created during data construction to keep training and rollout efficient. A naive scheme that always rewards visual assistance on positive-gain samples and penalizes it on negative-gain samples leads to overuse or over-suppression. Instead, we adopt a threshold-based scheme controlled by $\sigma$: only when the generated visual assistance exceeds $\sigma$ (in strength or proportion) do we assign a positive reward or penalty based on the offline visual-gain label $g \in \{-1, +1, 0\}$; boundary cases receive $r_e = 0$. The exact formulation is as follows:
\[ \small
r_e(o_i;g)=
\begin{cases}
0.2, & \text{if } g=+1 \ \land\ u(O) \le \sigma \ \land\ o_i=1,\\[2pt]
-0.2, & \text{if } g=-1 \ \land\ u(O) \ge \sigma \ \land\ o_i=1,\\[2pt]
0, & \text{otherwise},
\end{cases}
\]
where $o_i \in \{0, 1\}, i\in \{1, ...N\}$ means whether visual assistance are included in the i-th response, and $u(O) = \frac{1}{N}\sum_{i=1}^No_i$ indicates the proportion of $N$ responses that contain visual assistance.
This threshold design avoids indiscriminate visual assistance overuse or over-penalization.

%% file: sec/4_experiments.tex
\section{Experiments}
\label{sec:experiments}

\subsection{Experimental Setup}
In this section, we briefly introduce the implementation details, baselines, and evaluation settings.

\begin{table*}[t]
    \small
    \centering 
    \setlength{\tabcolsep}{5pt}
    \begin{tabular*}{\textwidth}{@{\extracolsep{\fill}}@{}l|cccccc@{}}
    \toprule
    \multirow{2}{*}{\textbf{Model}} & \multirow{2}{*}{\textbf{Average}} & \multicolumn{3}{c}{\textbf{Spatial Benchmarks}} & \multicolumn{2}{c}{\textbf{General Benchmarks}} \\
    \cmidrule{3-5}
    \cmidrule{6-7}
    ~ & ~ & \textbf{SIBench} & \textbf{Q-SpatialBench} &
    \textbf{MMVP} & \textbf{MMBench} & \textbf{MM-Vet} \\
    \midrule
    \multicolumn{7}{c}{\textit{Understanding-only Multimodal Large Language Models}} \\
    \midrule
    GPT-5 & 70.44 & 58.86 & 48.51 & 85.33 & 84.25 & 75.27 \\
    GPT-4o & 68.59 & 49.48 & 49.50 & 84.66 & 82.40 & 76.92 \\
    Qwen3VL-32B & 71.83 & 55.49 & 63.36 & 82.66 & 86.11 & 71.51 \\
    Qwen3VL-8B & 67.30 & 50.13 & 64.35 & 77.33 & 84.25 & 60.45 \\
    InternVL3.5-38B & 70.73 & 53.61 & 48.51 & 80.66 & 87.03 & 83.85 \\
    InternVL3.5-8B & 68.16 & 50.13 & 49.50 & 76.33 & 83.33 & 81.51 \\
    \midrule
    \multicolumn{7}{c}{\textit{Unified Multimodal Large Language Models}} \\
    \midrule
    Janus-Pro-7B & 53.45 & 42.73 & 53.46 & 62.33 & 60.18 & 48.53 \\
    Liquid-7B & 40.61 & 38.63 & 39.60 & 58.33 & 41.66 & 24.81 \\
    \midrule
    BAGEL & 60.49 & 43.57 & 47.52 & 74.33 & 72.22 & 64.81 \\
    BAGEL-PE & 59.45 & 48.25 & 51.48 & 72.66 & 69.44 & 55.41 \\
    BAGEL-RE & 62.24 & 43.98 & 48.51 & 75.33 & 77.77 & 65.59 \\
    \textbf{COOPER} & \textbf{66.42} & \textbf{50.07} & \textbf{57.42} & \textbf{78.66} & \textbf{80.55} & \textbf{65.41} \\
    \midrule
    \color{Red}\textbf{$\Delta$ (vs BAGEL)} & \color{Red}\textbf{+5.93} & \color{Red}\textbf{+6.50} & \color{Red}\textbf{+9.90} & \color{Red}\textbf{+4.33} & \color{Red}\textbf{+8.33} & \color{Red}\textbf{+0.60} \\
    \bottomrule
    \end{tabular*}
    \caption{\textbf{Main experimental results.} We evaluate COOPER on spatial reasoning/understanding benchmarks and general multimodal benchmarks. \textit{BAGEL-PE (Perception Enhancement)} is a variant that only learns auxiliary modality generation and, for each question, mechanically generates all auxiliary modalities before answering. \textit{BAGEL-RE (Reasoning Enhancement)} is a variant trained with the same RL data as COOPER but performs purely textual reasoning. Compared with the base model BAGEL, COOPER achieves an average improvement of \textbf{5.93\%}, and on distance and size estimation tasks it surpasses 38B open-source models and approaches proprietary models. Moreover, interleaved vision–language reasoning exhibits a higher performance ceiling than text-only reasoning.}
    \label{tab:main_results}
\end{table*}

\paragraph{Implementation details.}
We implement COOPER on top of BAGEL~\citep{deng2025emerging}.
(1) \textit{ Auxiliary Modality Generation}. For depth estimation, we use the synthetic indoor dataset Hypersim~\citep{roberts2021hypersim} and the outdoor Virtual KITTI~\citep{cabon2020virtual}, with far planes of 65m and 80m, respectively, and crop Virtual KITTI to the KITTI resolution~\citep{geiger2012we}. For segmentation, we adopt ADE20K~\citep{zhou2019semantic} with its original color palette. We jointly train depth and segmentation with a 1:1 sampling ratio and a learning rate of 5e-6 for one epoch.
(2) \textit{Adaptive Interleaved Reasoning}. After filtering (Section~\ref{sec:adaptive_reasoning}), we retain 17k samples and use GPT-4o to construct 7k SFT examples, training for one epoch with a learning rate of 5e-6.
On the remaining 10k samples, we apply GRPO with CPR reward threshold $\sigma=4$, KL coefficient $\beta=0.0$, learning rate 3e-6, batch size 128, and $N=8$ candidates for 20 steps until reward convergence. Training uses 8×NVIDIA H800 (80G) GPUs with random seed 42.
For evaluation, we adopt VLMEvalKit~\citep{duan2024vlmevalkit} and use DeepSeek-V3~\citep{liu2024deepseek} as an answer extractor to obtain final answers from model outputs.

\paragraph{Baseline approaches.}
We evaluate ten leading models to establish a strong baseline, including six MLLMs and three unified MLLMs. The MLLMs tested include open-source models InternVL3.5 (8B and 38B)~\citep{wang2025internvl3} and Qwen3VL (8B and 32B), as well as proprietary models GPT-4o, GPT-5. And the unified MLLMs include Janus-Pro~\citep{chen2025janus}, Liquid~\citep{wu2024liquid} and our base model BAGEL~\citep{deng2025emerging}.

\paragraph{Evaluation benchmarks.}
To assess effectiveness and generalization, we evaluate on three spatial perception/reasoning benchmarks and two general multimodal benchmarks:
SIBench~\citep{yu2025far}: curates nearly 20 open-source benchmarks covering 23 visual spatial reasoning settings. Since our method is trained on single-image data, we report results on SIBench’s single-image subset only.
Q-SpatialBench~\citep{liao2024reasoning}: comprises various size and distance estimation tasks to quantitatively measure fine-grained perception and estimation.
MMVP~\citep{tong2024eyes}: evaluates perception across nine different visual modes/patterns.
MMBench v1.1~\citep{liu2024mmbench}: an upgraded version of MMBench that removes low-quality items and adds harder ones.
MM-Vet~\citep{yu2023mm}: defines six core capabilities of multimodal models and poses complex questions that require combining multiple skills.

\subsection{Main Results}
Table~\ref{tab:main_results} presents our main experimental results, from which we can draw the following conclusions: 
\textbf{(1) COOPER significantly improves spatial understanding and reasoning.} Compared with unified MLLMs and the base model BAGEL, COOPER achieves an average gain of \textbf{6.91\%} on spatial understanding and reasoning. On Q-SpatialBench, it even surpasses proprietary GPT models and the 38B open-source InternVL3.5-38B, and reaches a performance level comparable to GPT-4o on the comprehensive spatial reasoning benchmark SIBench. 
\textbf{(2) While enhancing spatial reasoning, COOPER also improves general multimodal capability.} Relative to the base BAGEL model, COOPER yields an average improvement of 4.47\% on general multimodal benchmarks. 
\textbf{(3) Interleaved vision–language reasoning offers more headroom than text-only reasoning.} BAGEL-PE (Perception Enhancement) exhibits stronger spatial reasoning ability but weaker general-purpose problem-solving, whereas BAGEL-RE (Reasoning Enhancement) shows stronger general problem-solving ability but weaker spatial reasoning. In contrast, COOPER with interleaved reasoning achieves an average \textbf{4.59\%} improvement on spatial tasks over BAGEL-PE and an average \textbf{1.3\%} improvement on general tasks over BAGEL-RE.

\subsection{Variant Analysis}
\begin{table}[!t]
    \centering \small 
    \scalebox{0.99}{
    \begin{tabular*}{0.49 \textwidth}{@{\extracolsep{\fill}}@{}l|cccc@{}}
    \toprule
    \textbf{Variant} & \textbf{SIBench} &
    \textbf{Q-SpatialBench} & \textbf{MMBench} \\
    \midrule
    BAGEL & 43.57 & 47.52 & 72.22 \\
    + Stage1 & 43.82 & 55.44 & 76.85 \\
    + SFT & 48.06 & 54.35 & 78.48 \\
    + RL & 49.27 & 55.38 & 78.70 \\
    \; + $r_e$ & 50.07 & 57.42 & 80.55 \\
    \bottomrule
    \end{tabular*}
    }
    \caption{\textbf{Variant analysis.} Our analysis of COOPER variants shows that internalizing image generation further improves understanding-oriented performance, and that the visual-gain reward helps the model better select and allocate its capabilities.}
    \label{tab:variant_analysis}
\end{table}

\begin{figure*}
\centering
\includegraphics[width=1\textwidth]{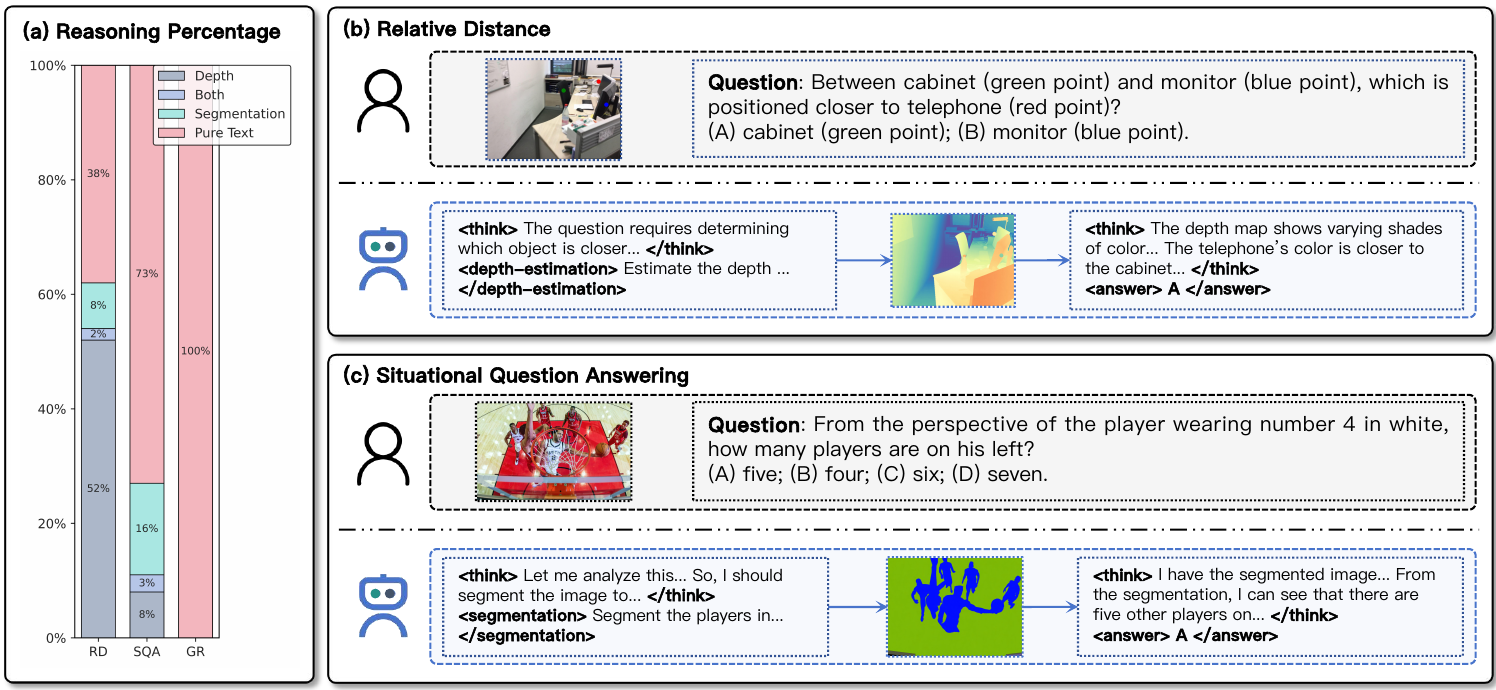}
\caption{\textbf{Reasoning Analysis.}
(a) COOPER adaptively selects its reasoning mode across tasks: for RD (Relative Distance) and SQA (Situational QA), it more often generates auxiliary multimodal signals, while for GR (Geometric Reasoning) it relies more on purely textual reasoning. (b) and (c) show how COOPER chooses to generate depth maps or highlight target objects in segmentation maps according to the task, thereby assisting its own reasoning. Additional reasoning and failure cases are provided in the \textit{supplementary materials}.
}
\label{fig:good_cases}
\end{figure*}

To better understand the contribution of each training stage in COOPER, we conduct ablation studies on two spatial benchmarks and one general benchmark. From the results in Table~\ref{tab:variant_analysis}, we draw the following key observations: 
\textbf{(1) Internalizing generative capabilities significantly improves task-specific understanding.} After injecting depth estimation and segmentation as internal generative skills (i.e., full generative-task training), the model achieves a substantial 7.92\% gain on Q-SpatialBench, which focuses on distance and size estimation, reaching performance close to the final SFT+RL model. It also yields about 4.63\% improvement on general benchmarks. 
\textbf{(2) For a relatively weak base model, SFT plays a crucial role.} The results show that SFT alone brings an average improvement of 5.86\%, and subsequent RL adds a further 6.68\% gain. This suggests that, for base models like BAGEL with limited initial capability, SFT already provides substantial improvements, while RL primarily serves as a refinement step on top of SFT. 
\textbf{(3) The exploration-guided reward $r_e$ further enhances capability selection and scheduling.} Compared with RL that uses only format and outcome rewards, adding the visual-gain reward consistently improves performance across three benchmarks, indicating that it helps the model decide when and how strongly to invoke visual assistance.

\subsection{Further Analysis}

\paragraph{Reasoning analysis.}
To better understand COOPER’s reasoning patterns, we perform quantitative and qualitative analyses in Figure~\ref{fig:good_cases}. Overall, \textbf{COOPER adaptively selects auxiliary modalities according to the task:}
(1) From the quantitative analysis in Figure~\ref{fig:good_cases}~(a), we observe that COOPER prefers different auxiliary modalities for different task types. For RD (Relative Distance) tasks, the model tends to generate depth maps; for SQA (situational QA) tasks, it more often highlights target objects in segmentation maps to assist reasoning. In contrast, for GR (Geometric Reasoning) tasks the model mainly relies on pure textual reasoning, since additional auxiliary modalities provide almost no benefit in this setting.
(2) The qualitative analysis in Figure~\ref{fig:good_cases}~(b) and (c) further illustrates reasoning examples for RD and SQA tasks. COOPER first analyzes the task, then selects an appropriate auxiliary modality, and finally combines information from the original image and the generated auxiliary modality to produce the final answer. 
More reasoning examples and failure cases can be found in the \textit{supplementary materials}.

\begin{figure}
\centering
\includegraphics[width=0.48 \textwidth]{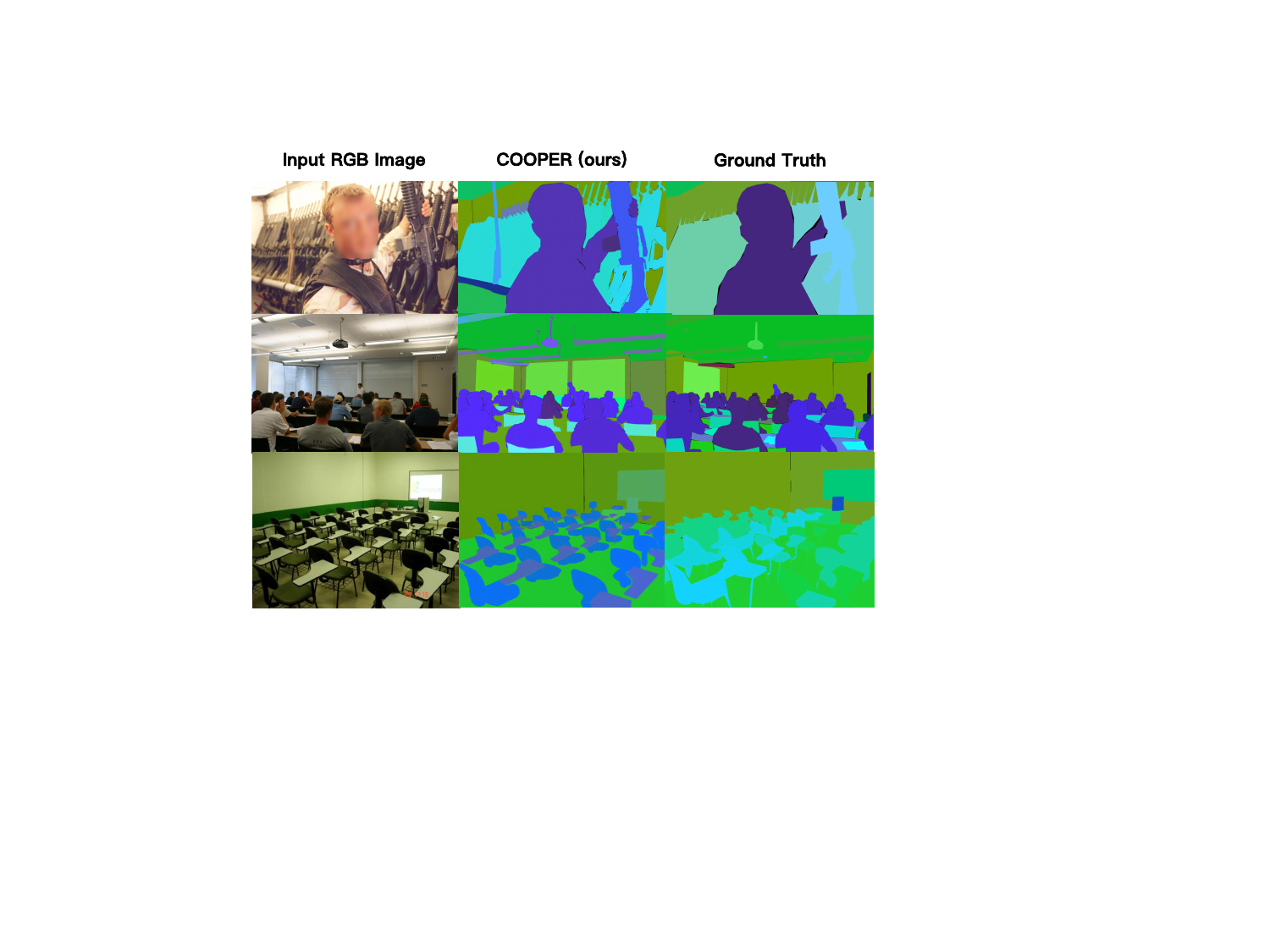}
\caption{\textbf{Segmentation Cases.}
Qualitative comparison between the COOPER and the ground-truth segmentation maps.
} 
\label{fig:seg_comp_mini}
\end{figure}

\begin{figure}
\centering
\includegraphics[width=0.48 \textwidth]{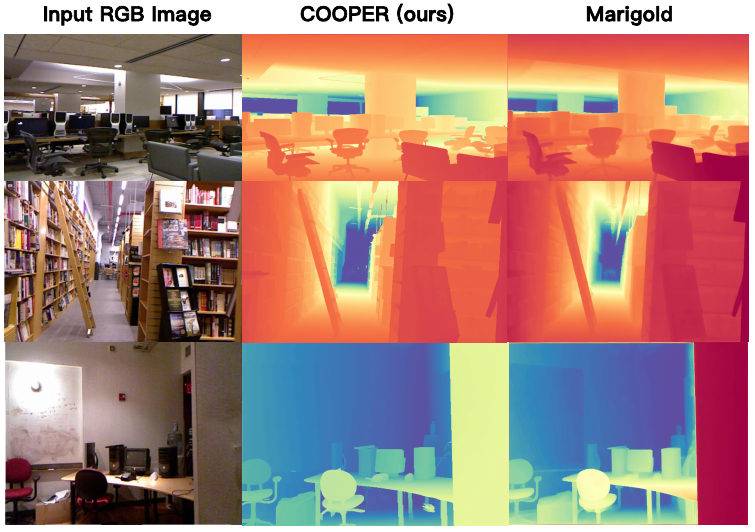}
\caption{\textbf{Depth estimation cases.}
Qualitative comparison between COOPER’s depth maps and the Marigold depth maps.
} 
\label{fig:depth_comp_mini}
\end{figure}

\begin{table}[!t]
    \centering \small 
    \begin{tabular*}{0.3 \textwidth}{@{\extracolsep{\fill}}@{}l|cc@{}}
    \toprule
    \multirow{2}{*}{Models} & \multicolumn{2}{c}{\textbf{NYUv2}~\citep{silberman2012indoor}} \\
    \cmidrule{2-3}
    ~ & AbsRel $\downarrow$ & $\delta1\uparrow$ \\
    \midrule
    DPT~\citep{ranftl2021vision} & 9.8 & 90.3 \\
    Marigold~\citep{ke2024repurposing} & \underline{5.5} & \textbf{96.4} \\
    COOPER & \textbf{0.5} & \underline{93.2} \\
    \bottomrule
    \end{tabular*}
    \caption{\textbf{Quantitative analysis of depth estimation.}
    COOPER’s performance on out-of-domain depth benchmarks is comparable to that of dedicated depth estimation models.
    }
    \label{tab:depth}
\end{table}

\paragraph{Auxiliary modality analysis.}
We provide qualitative and quantitative evidence that \textbf{COOPER’s auxiliary modality generation is comparable to specialized models}.
\textit{(1) Segmentation}: Because segmentation labels are mapped to RGB values, standard segmentation metrics are hard to apply. We only provide the qualitative comparison of segmentation result (Figure~\ref{fig:seg_comp_mini}). COOPER often yields finer boundaries and more distinguishable colors than the ground truth—for example, capturing the precise shape of the gun barrel rather than a coarse triangular region, and assigning desk regions colors that stand out more clearly from the background.
\textit{(2) Depth estimation}: Here we compare with Marigold, as NYUv2 ground-truth depth maps are hard for humans to distinguish. 
On the out-of-domain NYUv2 dataset, COOPER produces sharper depth maps with clearer boundaries than the specialized model Marigold (Figure~\ref{fig:depth_comp_mini}).
Quantitatively (Table~\ref{tab:depth}), COOPER attains AbsRel and $\delta1$ performance on par with Marigold. More examples are provided in the \textit{supplementary materials}.

%% file: sec/5_related.tex
\section{Related Work}
\label{sec:related}

\subsection{Multimodal Chain-of-Thought}
Chain-of-Thought (CoT)~\citep{wei2022chain} improves complex problem solving by decomposing tasks into textual reasoning steps, but in multimodal settings text alone is insufficient—humans naturally rely on visual aids such as sketches and auxiliary lines. Multimodal CoT~\citep{lin2025mind, wang2025multimodal, su2025thinking} follows this idea by integrating visual information into the reasoning process. Early work depends on external tools or expert models~\citep{su2025pixel, zheng2025deepeyes, wu2025reinforcing, zhou2025reinforced, zhang2025thyme, hu2024visual}, resulting in template-like and inflexible visual reasoning. More recent approaches use unified MLLMs’ image–text understanding and generation to elicit intrinsic multimodal CoT~\citep{yang2025machine, li2025imagine, gu2025thinkmorph, shi2025mathcanvas, li2025zebra}, showing benefits in math and perception tasks via visual cues. However, these methods mainly modify natural images and rarely treat outputs of visual understanding tasks (e.g., depth, segmentation) as intermediate reasoning states, even though such signals are crucial for visual spatial reasoning. In this paper, we equip models with these visual understanding abilities and enable them to invoke them dynamically during inference, yielding a more powerful multimodal CoT for spatial reasoning.

\subsection{Visual Spatial Reasoning}
Visual Spatial Reasoning~\citep{yang2025thinking} requires models not only to understand semantics and localize targets, but also to reason about spatial relationships and infer 3D structure from 2D images. Existing work mainly follows two directions:
(1) Perception enhancement~\citep{cheng2024spatialrgpt, zhou2025roborefer, liu2025ssr, wang2025spatialclip, ma2025spatialllm, feng2025towards, fan2025vlm, wu2025spatial, zheng2025learning}, which introduces auxiliary modalities (e.g., depth, segmentation) to improve 3D perception from 2D inputs;
(2) Reasoning enhancement~\citep{ray2024sat, liu2025spatialcot, cai2025spatialbot, ai2025m2, ouyang2025spacer, ma2025spatialreasoner, liao2025improved, wang2025svqa, li2025videochat}, which borrows techniques from text-based reasoning to strengthen spatial reasoning.
Although perception and reasoning are tightly coupled, prior work typically improves them in isolation. In this paper, we leverage unified MLLMs’ strong image–text understanding and generation capabilities to integrate auxiliary modalities into a multimodal chain-of-thought, allowing the model to decide during inference when to invoke perception or reasoning enhancement for more comprehensive visual spatial reasoning~\footnote{More related works, future directions, and limitations are provided in the supplementary materials.}.

%% file: sec/6_conclusion.tex
\section{Conclusion}
\label{sec:conclusion}
In this work, we aim to build a stronger visual spatial reasoning model by dynamically combining perception enhancement and reasoning enhancement within an intrinsic multimodal chain-of-thought, and we introduce COOPER to this end. Concretely, starting from a unified MLLM, we first inject the necessary perception capabilities by enabling the model to generate depth maps and segmentation maps. We then carefully curate datasets for both SFT and RL: supervised fine-tuning is used to endow the model with basic dynamic activation behavior, and reinforcement learning with CPR reward is further applied to strengthen its reasoning ability and generalization.

%% file: sec/X_suppl.tex
\clearpage
\setcounter{page}{1}
\maketitlesupplementary

\section{Additional Related Works}
\subsection{Unified Multimodal Large Language Models}
Unified MLLMs~\citep{zhang2025unified} aim to build a single architecture that can understand and generate data across multiple modalities. They process diverse inputs (e.g., text, images, video, audio) and produce outputs in one or more modalities within a unified framework. Such architectures typically consist of three core components: modality-specific encoders that map inputs into a shared representation space, a multimodal fusion backbone for cross-modal reasoning, and modality-specific decoders for tasks like text or image generation.
Early work focused on fully unified autoregressive models that discretize visual information into tokens via vector quantization~\citep{van2017neural} and train the model with a unified next-token prediction objective~\citep{team2024chameleon, chern2024anole, wang2024emu3, xie2024show, wu2024liquid, wu2024vila}. 
Subsequent studies, however, observed that such visual tokenization tends to lose substantial fine-grained visual details. To address this, starting from Janus~\citep{wu2025janus}, visual encoding has been decoupled into two branches: one dedicated to visual understanding and the other to visual generation. The former uses semantic encoders such as CLIP-ViT to capture visual semantics, while the latter employs Diffusion VAEs to encode and decode generative visual information~\citep{chen2025janus, xiao2025mindomni, xiao2025omnigen, deng2025emerging}. Among these, BAGEL is pretrained on large-scale image–text interleaved data and exhibits strong cross-modal understanding and generation capabilities, so we adopt BAGEL as the base model in this work.

\subsection{Unified Diffusion Models}
Diffusion models~\citep{lipman2022flow, albergo2022building, liu2022flow, peebles2023scalable} have achieved remarkable success in image and video generation in recent years. The core idea is to gradually add noise to the data and train a model to reverse this process, denoising step by step to produce the final output.
Building on diffusion models pretrained on large-scale web data, recent works have begun to explore whether this powerful foundation can be leveraged to construct a unified model that handles a wide range of image-to-image tasks, such as depth estimation, image segmentation, optical flow estimation, deraining, dehazing, and deblurring. Early diffusion-based approaches, such as VPD~\citep{zhao2023unleashing} and DDP~\citep{ji2023ddp}, typically attach task-specific decoders on top of a diffusion backbone and train a separate diffusion model for each task. Later, Marigold~\citep{ke2024repurposing} demonstrated that even using the original diffusion architecture and modeling paradigm alone can already match or surpass specialized perception models on depth estimation.
More recently, models~\citep{xu2025jodi} such as OneDiffusion~\citep{le2025one}, DICEPTION~\citep{zhao2025diception}, and Qwen-Image~\citep{wu2025qwen} have showcased even stronger unification capabilities. 
Importantly, the outputs of these visual understanding tasks—especially depth estimation and segmentation—are highly beneficial for spatial reasoning. However, these explorations are largely confined to the diffusion backbone itself. In this work, we take a further step by injecting these capabilities into unified MLLMs~\citep{deng2025emerging} and leveraging its strong image and text generation abilities to explicitly incorporate auxiliary modality generation as part of a multi-modal chain-of-thought, thereby enhancing the model’s spatial reasoning capability.

\section{Future Directions}
\subsection{Joint Text-Image GRPO for Multimodal CoT Reasoning}
During the reinforcement learning stage, we currently adopt a standard GRPO objective whose reward is applied only to generated text sequences, and thus cannot directly optimize the model’s image generation behavior. This text-centric optimization limits, to some extent, the ability of unified MLLMs to acquire intrinsic multimodal chain-of-thought (CoT) reasoning. Recently, however, GRPO variants tailored for image generation have been proposed, enabling models to optimize their visual outputs according to task-specific rewards~\citep{liu2025flow, wang2025pref, he2025tempflow, yuan2025ar, wu2025rewarddance}. Building on this, a promising direction is to integrate text reasoning rewards and image generation rewards within a unified GRPO framework: encouraging more accurate and interpretable CoT reasoning on the textual side, while simultaneously promoting visual outputs that are better aligned with the reasoning process and task objectives. We expect that such a multimodal joint optimization strategy can further enhance the multimodal CoT reasoning capabilities of unified MLLMs.

\subsection{Toward Efficient Unified MLLMs for Long-Video Spatial Reasoning}
Constrained by the architecture and inference efficiency of the BAGEL model (e.g., its current incompatibility with vLLM and other inference acceleration frameworks), our experiments are presently limited to single-image spatial reasoning tasks. However, real-world spatial reasoning applications typically require handling long-horizon, continuous video streams, which presents a substantial gap between our current experimental setup and practical deployment scenarios. 
Although several contemporaneous efforts aim to improve BAGEL’s inference efficiency, for example by adapting it to vLLM~\footnote{https://github.com/vllm-project/vllm/issues/18793} or adopting more efficient architectures~\citep{lu2025hyper}, these approaches are either still under development or exhibit a noticeable performance drop compared with the original BAGEL model, leaving them far from being ready for real-world use.
At the same time, our study shows that intrinsic multimodal CoT exhibits greater potential than purely text-based reasoning models. Motivated by this, an important future direction is to design unified MLLMs that simultaneously achieve high inference efficiency and strong intrinsic multimodal CoT capabilities, enabling them to scale to long-video, spatial reasoning tasks that more closely reflect real-world application scenarios while maintaining high reasoning quality.

\subsection{Enriching Auxiliary Modalities in Multimodal CoT for Spatial Reasoning}
In our experimental setup, in order to simplify the problem, we currently consider only two forms of auxiliary modalities: depth estimation, which provides geometric information, and segmentation, which provides semantic information. However, richer auxiliary modalities, such as 3D point cloud data and even natural images from real-world scenes, are also expected to provide beneficial support for spatial reasoning~\citep{li2025imagine, gu2025thinkmorph, yang2025machine, wu2025reinforcing}. Building on this, future experimental studies and real-world application systems can further incorporate a broader range of auxiliary modalities into the intrinsic multimodal CoT framework, helping the model acquire stronger spatial reasoning capabilities.

\section{Additional Experimental Details and Qualitative Examples}
\begin{figure*}
\centering
\includegraphics[width=0.8 \textwidth]{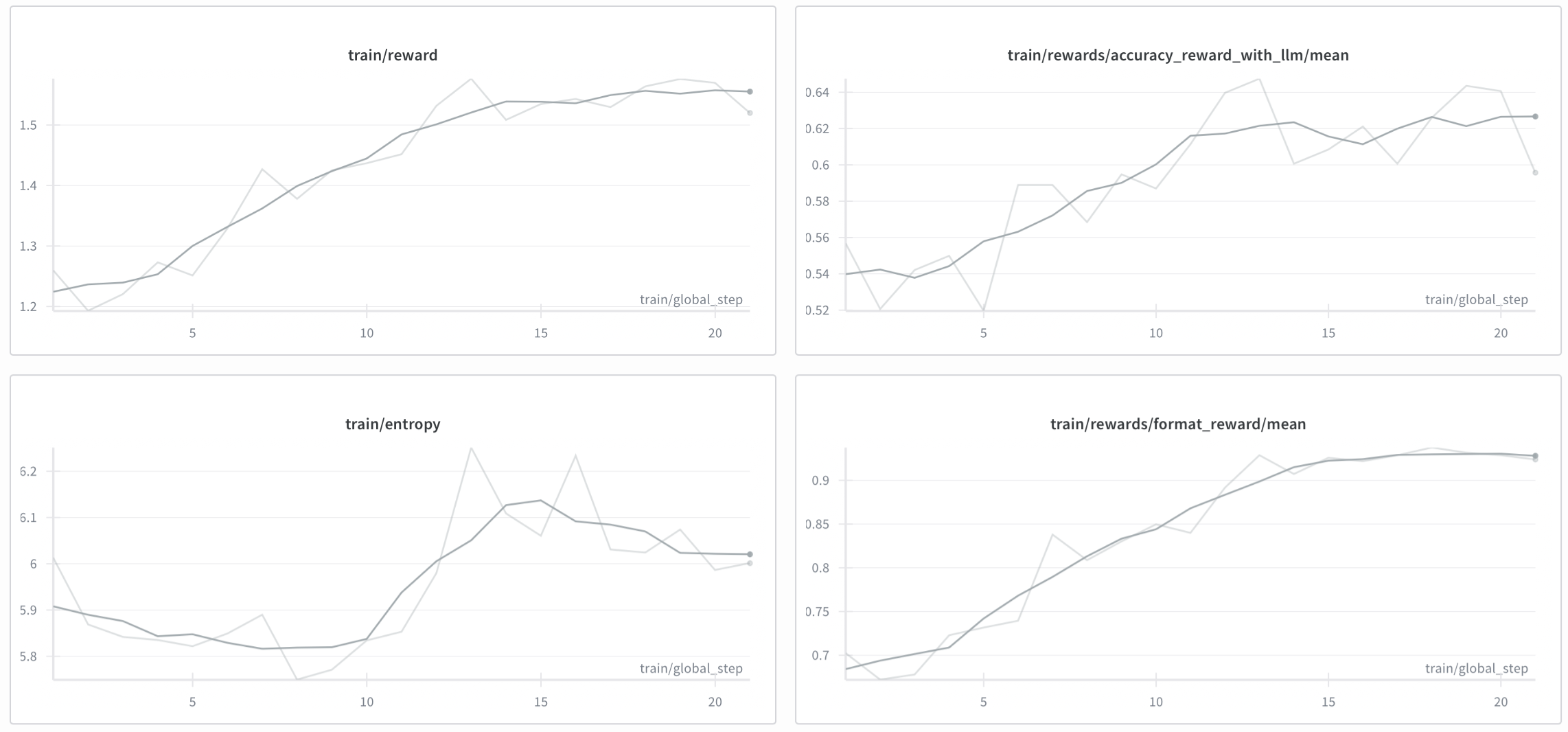}
\caption{\textbf{Reward curve.}
The reward curve of COOPER during training.
} 
\label{fig:reward}
\end{figure*}

\begin{figure*}
\centering
\includegraphics[width=0.85\textwidth]{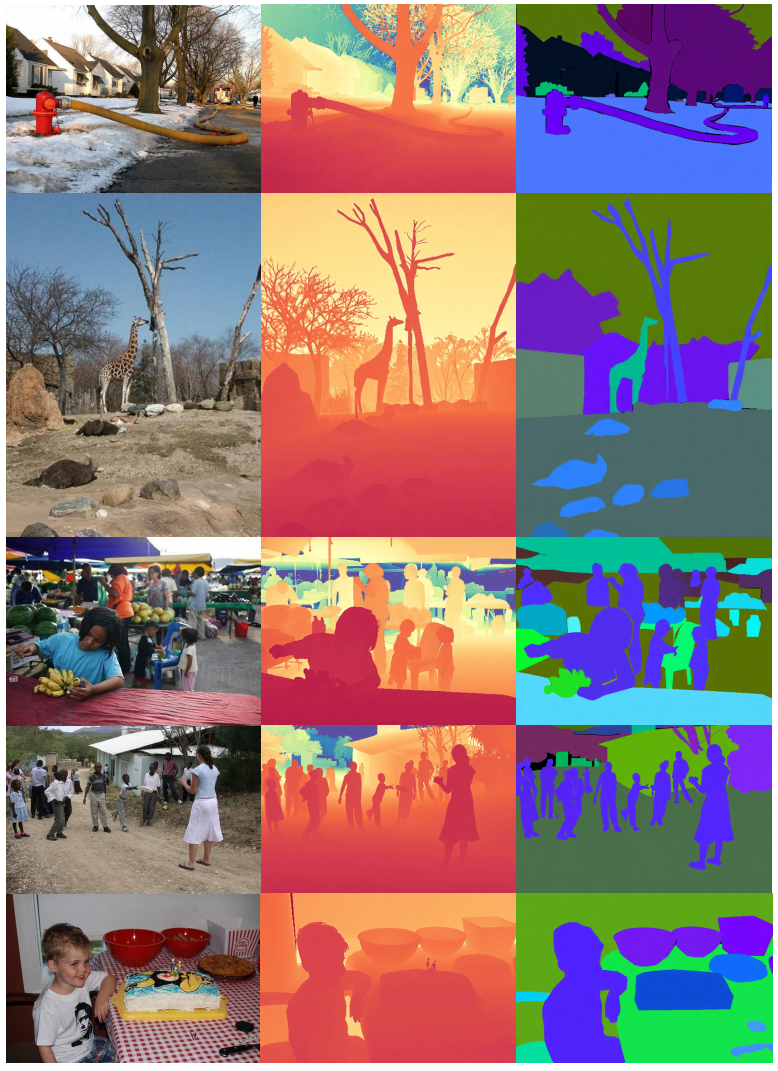}
\caption{\textbf{Generation cases from COOPER.}
The figure illustrates that COOPER generates different auxiliary modalities for the same input image.
} 
\label{fig:generation_cases}
\end{figure*}

\begin{figure*}
\centering
\includegraphics[width=0.85\textwidth]{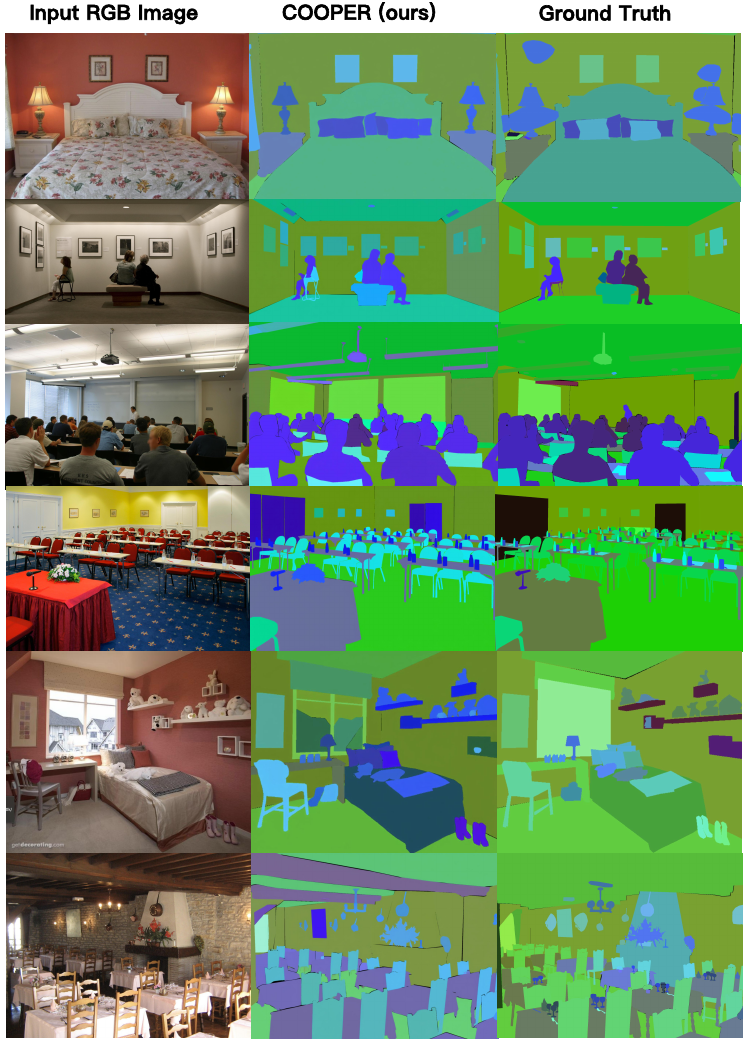}
\caption{\textbf{Segmentation Cases.}
Qualitative comparison between the COOPER and the ground-truth segmentation maps.
} 
\label{fig:seg_comp}
\end{figure*}

\begin{figure*}
\centering
\includegraphics[width=0.85\textwidth]{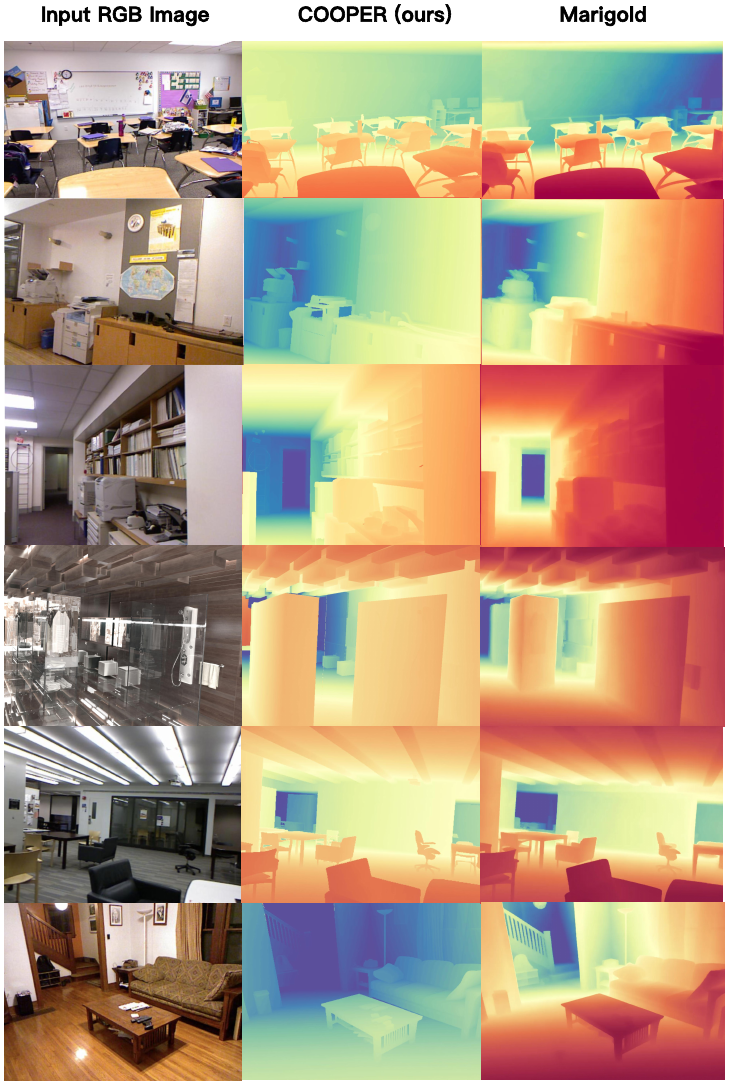}
\caption{\textbf{Depth estimation cases.} 
Qualitative comparison between COOPER’s depth maps and the Marigold depth maps.
} 
\label{fig:depth_comp}
\end{figure*}

\begin{figure*}
\centering
\includegraphics[width=0.85\textwidth]{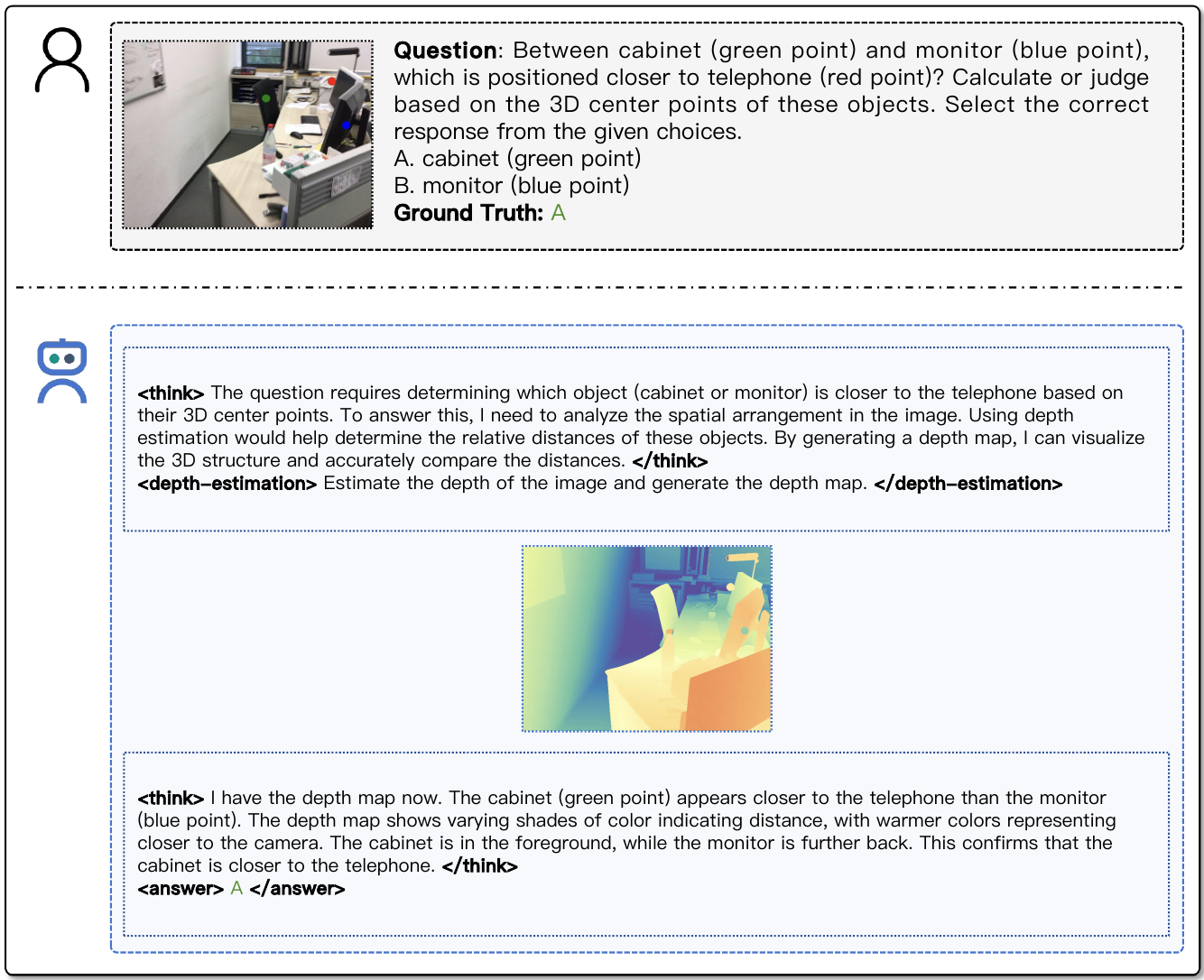}
\caption{\textbf{Reasoning cases 1.} 
An example of depth-estimation–enhanced reasoning in COOPER.
} 
\label{fig:depth_good_0}
\end{figure*}

\begin{figure*}
\centering
\includegraphics[width=0.85\textwidth]{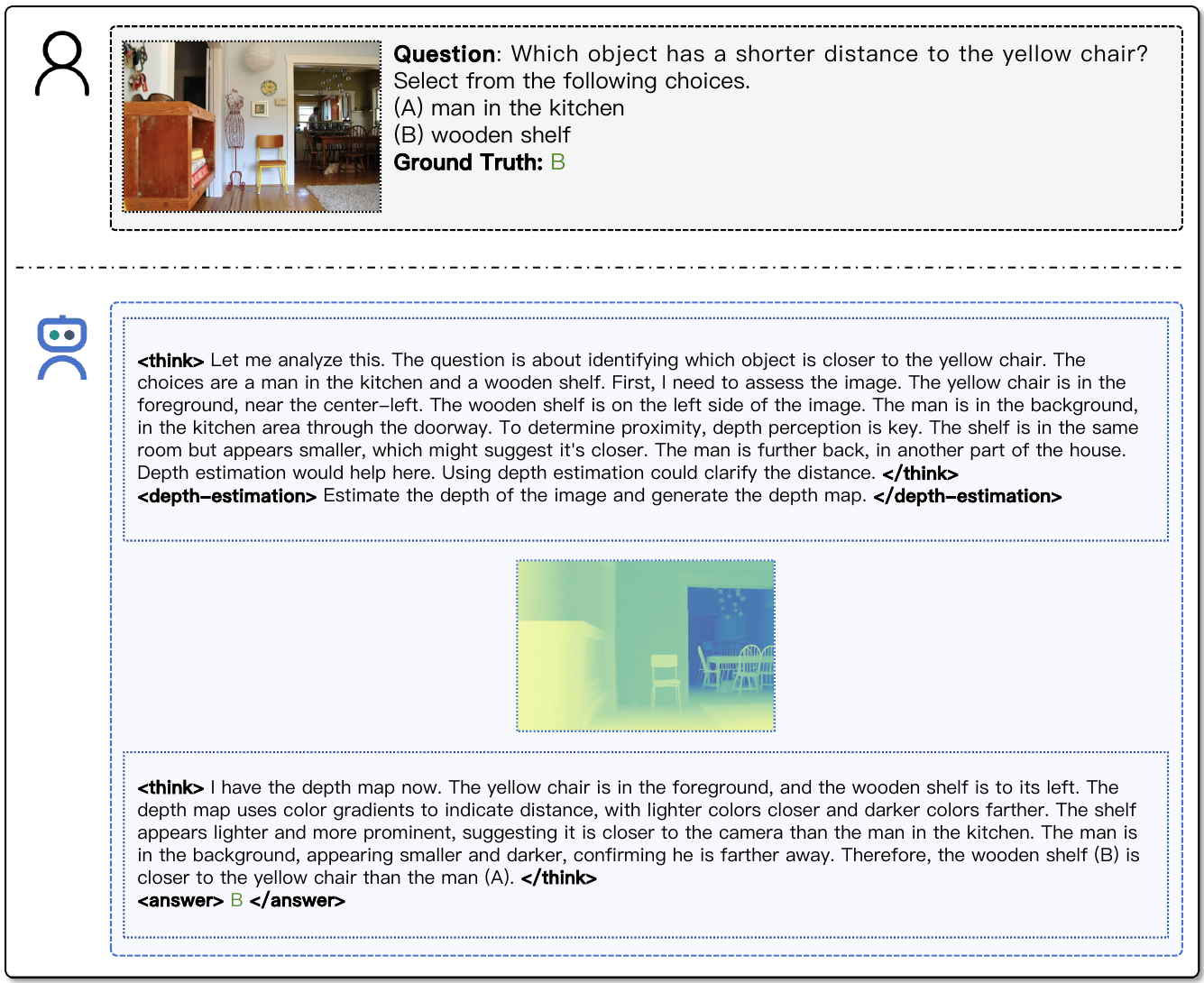}
\caption{\textbf{Reasoning cases 2.} 
An example of depth-estimation–enhanced reasoning in COOPER.
} 
\label{fig:depth_good_1}
\end{figure*}

\begin{figure*}
\centering
\includegraphics[width=0.85\textwidth]{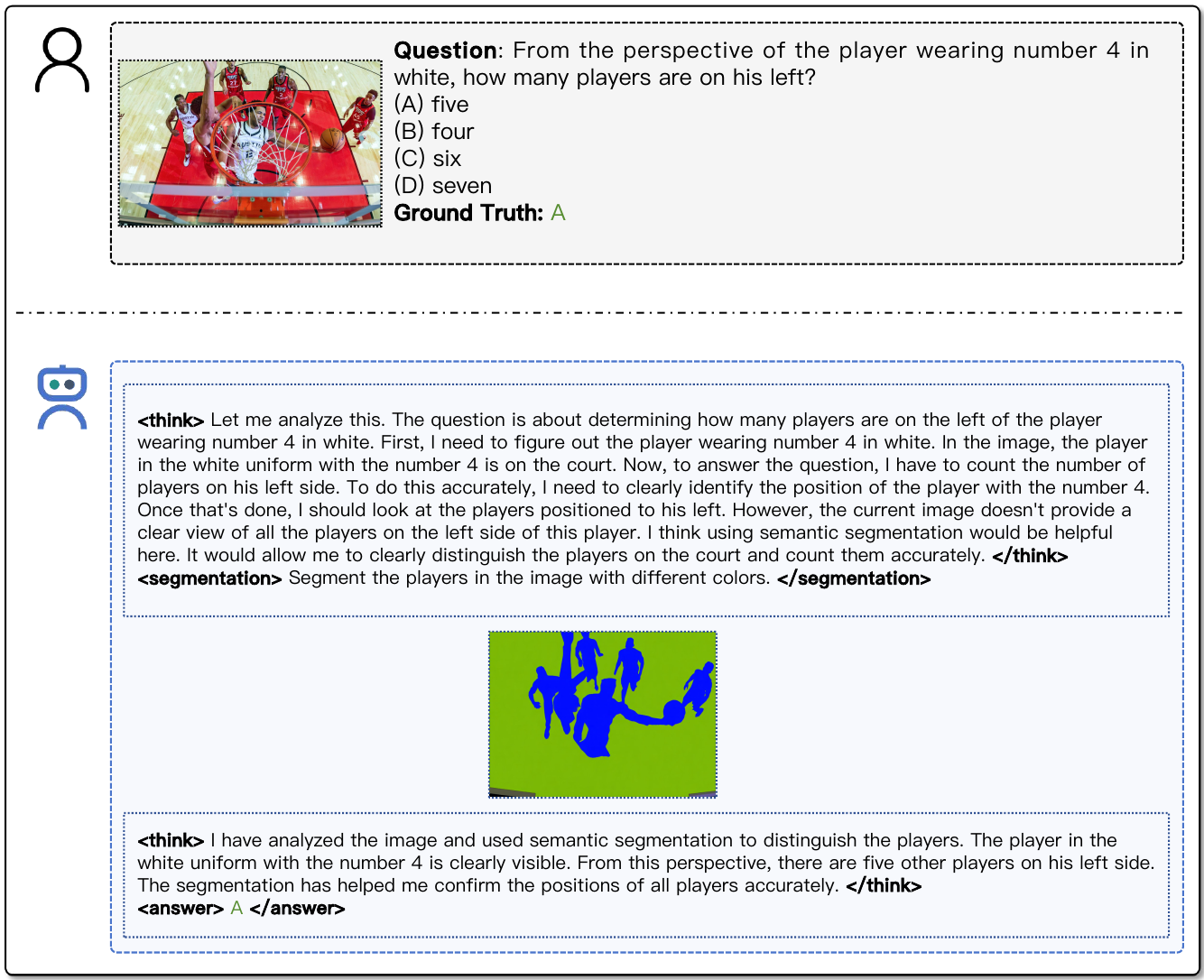}
\caption{\textbf{Reasoning cases 3.} 
An example of segmentation–enhanced reasoning in COOPER.
} 
\label{fig:seg_good_0}
\end{figure*}

\begin{figure*}
\centering
\includegraphics[width=0.85\textwidth]{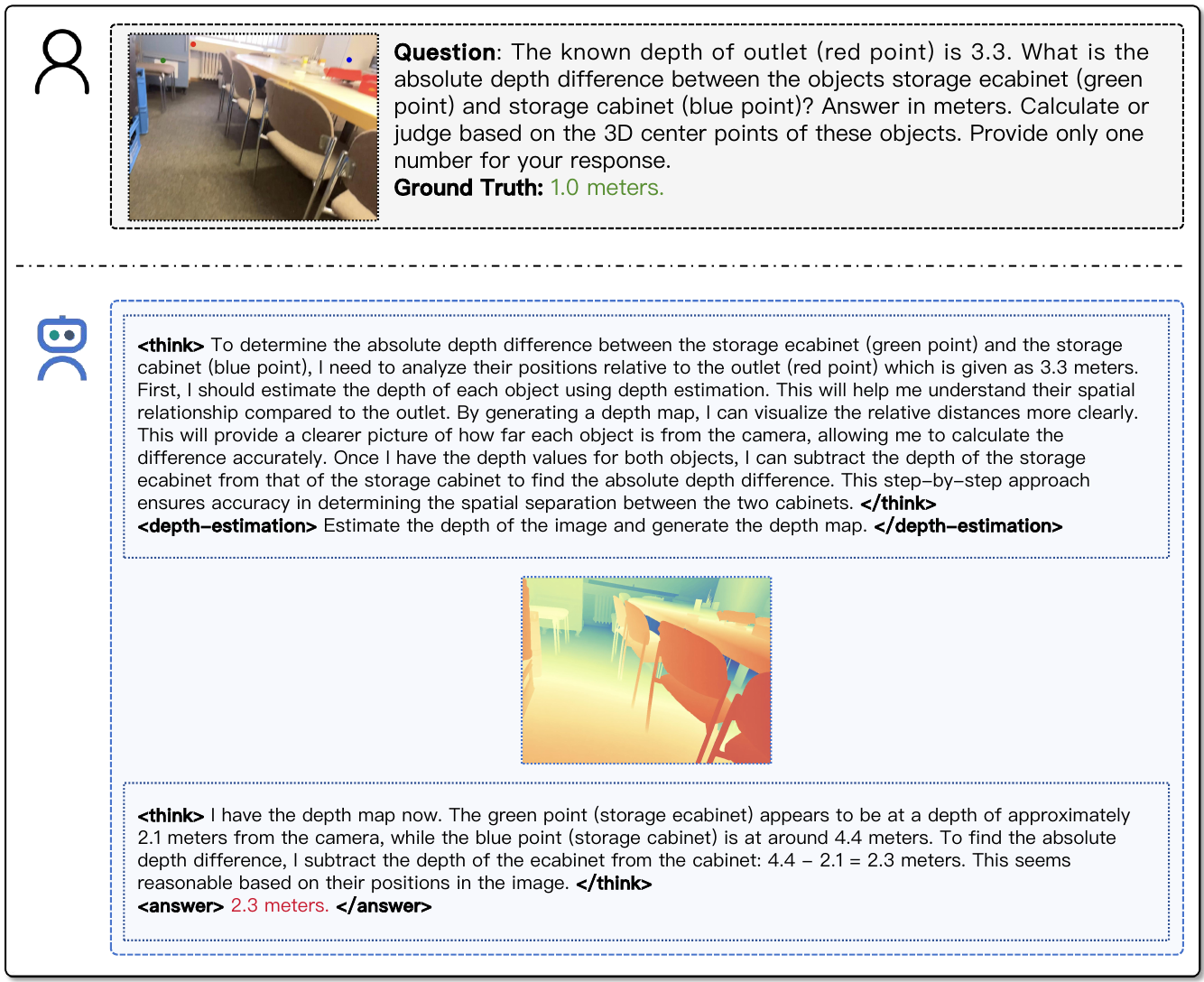}
\caption{\textbf{Reasoning cases 4.} 
A failure example of depth-estimation–enhanced reasoning in COOPER.
} 
\label{fig:depth_bad_0}
\end{figure*}

In this section, we present more experimental details and a richer set of reasoning and generation cases. 
\paragraph{Experimental details.}
For the generation cases, we use $50$ denoising steps to generate the auxiliary modalities, whereas for the \textit{Marigold} results we set the ensemble size to $10$ and use $20$ denoising steps to generate the depth maps.
Figure~\ref{fig:reward} shows the evolution of rewards and entropy during COOPER’s training. The overall reward steadily increases and begins to converge after roughly 15 training steps. Both the accuracy reward and the format reward improve consistently over time, indicating that the model keeps getting better in terms of answer correctness and output formatting. Meanwhile, the entropy of COOPER’s policy first decreases and then rises, before slightly dropping and stabilizing, suggesting that the model gradually develops a more stable policy while still maintaining a certain level of exploration.
\paragraph{More Cases.}
Figure~\ref{fig:generation_cases} illustrates how COOPER generates different types of auxiliary modalities from the same input image, while Figure~\ref{fig:seg_comp} and Figure~\ref{fig:depth_comp} provide additional examples of segmentation and depth estimation, respectively. Figure~\ref{fig:depth_good_0}, Figure~\ref{fig:depth_good_1}, Figure~\ref{fig:seg_good_0}, Figure~\ref{fig:depth_bad_0} showcase more complete reasoning trajectories. In particular, Figure~\ref{fig:depth_good_0} and Figure~\ref{fig:depth_good_1} show successful uses of COOPER for depth estimation: the model first analyzes the question, then generates a depth map as an auxiliary modality, and finally combines the depth map with the original image to answer the question. Figure~\ref{fig:seg_good_0} demonstrates a similar reasoning process when using segmentation. In contrast, Figure~\ref{fig:depth_bad_0} presents a failure case: although COOPER correctly selects a depth map as the auxiliary modality, it hallucinates the distance between two objects during reasoning, leading to an incorrect final answer.

\section{Prompt Templates}

\begin{figure*}
\centering
\includegraphics[width=0.85\textwidth]{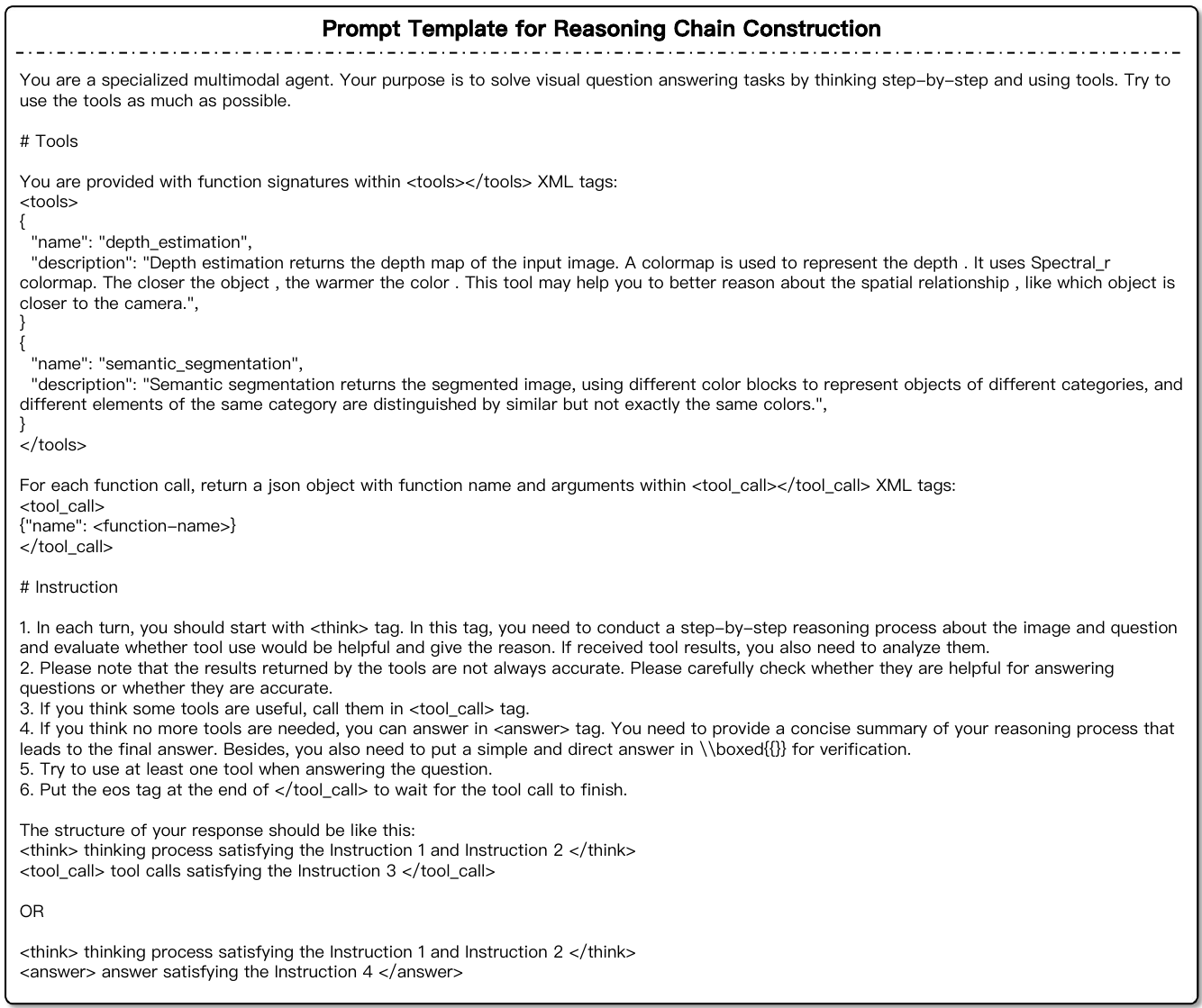}
\caption{\textbf{Prompt template for reasoning chain construction.}
} 
\label{fig:reason_chain_prompt}
\end{figure*}

\begin{figure*}
\centering
\includegraphics[width=0.85\textwidth]{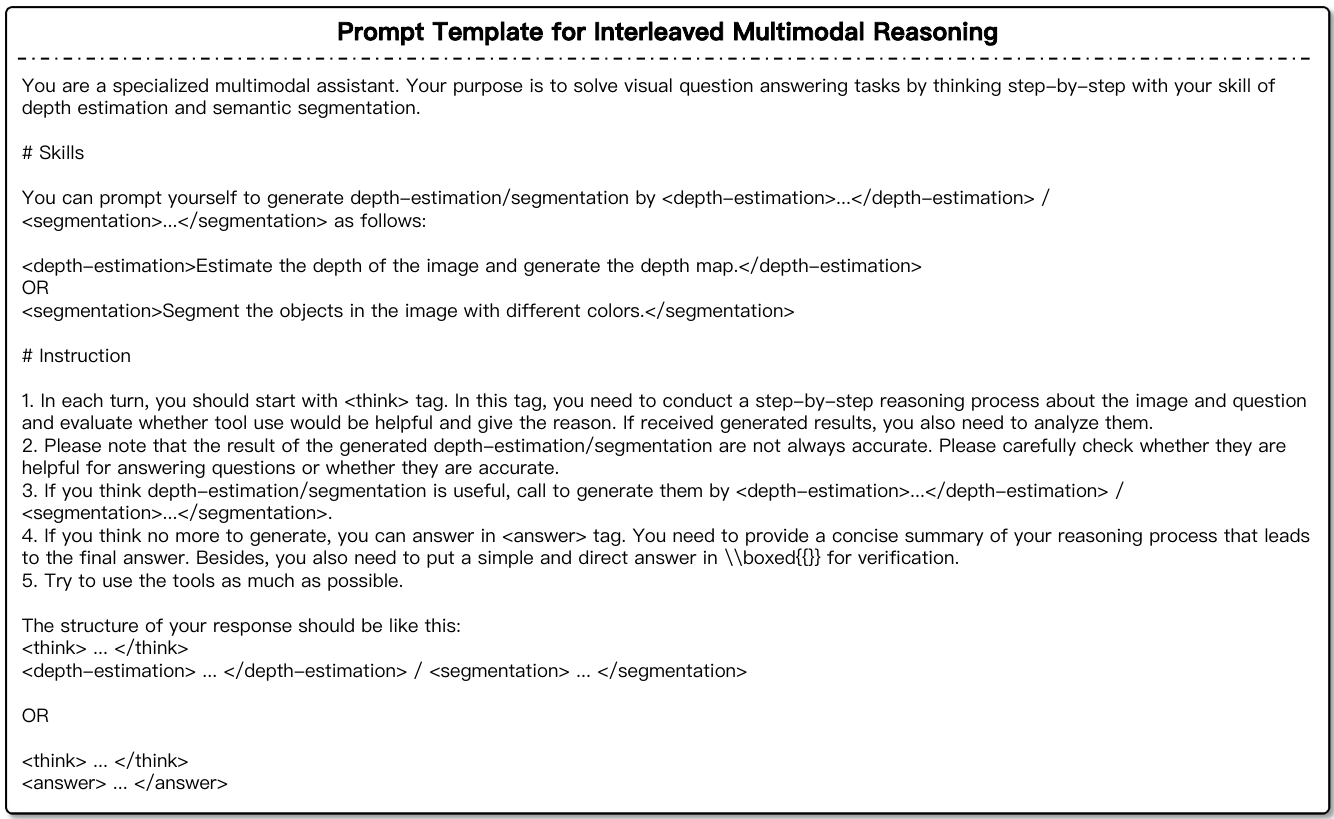}
\caption{\textbf{Prompt template for interleaved multimodal reasoning in COOPER.}
} 
\label{fig:interleaved_reason_prompt}
\end{figure*}

\begin{figure*}
\centering
\includegraphics[width=0.85\textwidth]{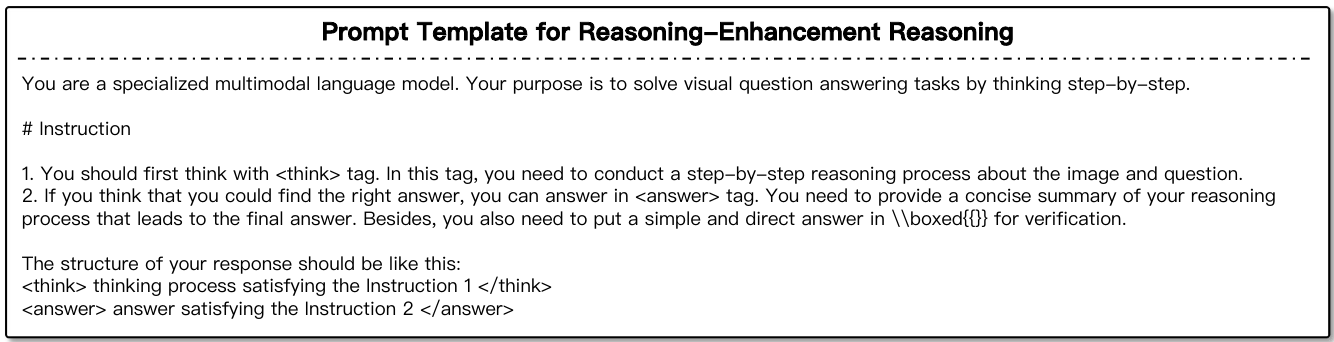}
\caption{\textbf{Prompt template for reasoning-enhancement reasoning.}
} 
\label{fig:reason_enhancement_prompt}
\end{figure*}

\begin{figure*}
\centering
\includegraphics[width=0.85\textwidth]{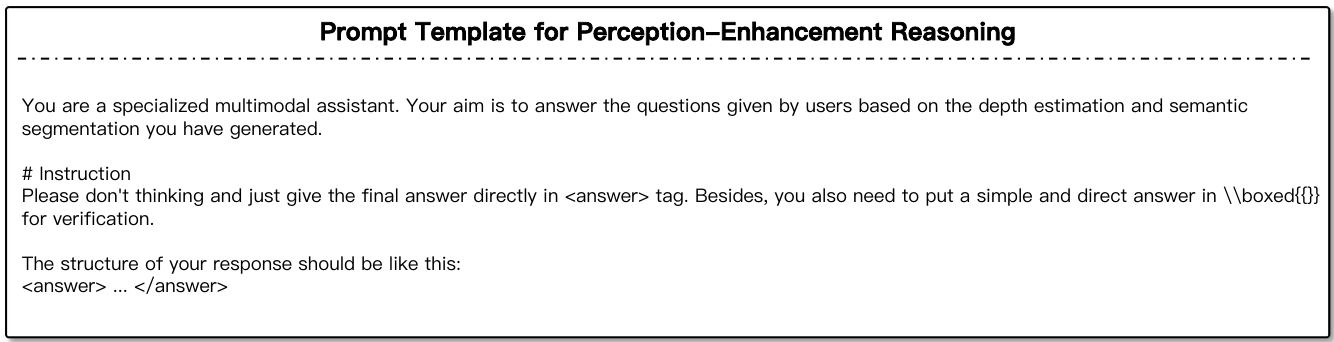}
\caption{\textbf{Prompt template for perception-enhancement reasoning.}
} 
\label{fig:perception_enhancement_prompt}
\end{figure*}

In this section, we present all the prompt templates used in our experiments. Figure~\ref{fig:reason_chain_prompt} shows the prompt used with GPT to construct reasoning chains, while Figure~\ref{fig:interleaved_reason_prompt} illustrates the prompt used by COOPER for interleaved multimodal reasoning. Figure~\ref{fig:reason_enhancement_prompt} and Figure~\ref{fig:perception_enhancement_prompt} respectively present the prompt templates used in the reasoning-enhancement and perception-enhancement experiments.